\def\year{2019}\relax
\documentclass[letterpaper]{article} 
\usepackage{aaai19}  
\usepackage{times}  
\usepackage{helvet}  
\usepackage{courier}  
\usepackage{url}  
\usepackage{graphicx}  

\usepackage{graphicx}
\usepackage{subcaption}
\usepackage{enumitem}
\usepackage{amsmath}
\usepackage{booktabs}
\usepackage{footnote}
\usepackage{cleveref}
\usepackage{adjustbox}
\usepackage{bm}
\usepackage{array}
\usepackage{times}
\usepackage{latexsym}
\usepackage{url}
\usepackage{natbib}
\usepackage{multirow}
\DeclareMathOperator*{\argmax}{argmax}

\title{Syntax-aware Neural Semantic Role Labeling\thanks{Zhenghua Li is the corresponding Author.}}
\author{Qingrong Xia$^1$, Zhenghua Li$^1$, Min Zhang$^1$, Meishan Zhang$^2$, Guohong Fu$^2$, Rui Wang$^3$, Luo Si$^3$ \\
$^1$Institute of Artificial Intelligence, School of Computer Science and Technology, Soochow University, China \\
$^2$School of Computer Science and Technology, Heilongjiang University, China, $^3$Alibaba Group, China \\
$^1$kirosummer.nlp@gmail.com, \{zhli13, minzhang\}@suda.edu.cn \\
$^2$mason.zms@gmail.com, ghfu@hotmail.com, $^3$\{masi.wr, luo.si\}@alibaba-inc.com
}

\begin{document}
\maketitle

\begin{abstract}
Semantic role labeling (SRL), also known as shallow semantic parsing, is an important yet challenging  task in NLP. 
Motivated by the close correlation between syntactic and semantic structures, traditional discrete-feature-based SRL approaches make heavy use of syntactic features. 
In contrast, 
deep-neural-network-based approaches usually encode the input sentence as a word sequence without considering the syntactic structures. 
In this work, we investigate several previous approaches for encoding syntactic trees, and make a thorough study on whether extra syntax-aware representations are beneficial for neural SRL models.
Experiments on the benchmark CoNLL-2005 dataset show that syntax-aware SRL approaches can effectively improve performance over a strong baseline with external word representations from ELMo.
With the extra syntax-aware representations, our approaches achieve new state-of-the-art 85.6 F1 (single model) and 86.6 F1 (ensemble) on the test data, outperforming the corresponding strong baselines with ELMo by 0.8 and 1.0, respectively.
Detailed error analysis are conducted to gain more insights on the investigated approaches. 

\end{abstract}

\section{Introduction}

Semantic role labeling (SRL), also known as shallow semantic parsing, is an important yet challenging task in NLP.   
Given an input sentence and one or more predicates, SRL aims to determine the semantic roles of each predicate, i.e., \underline{who} did \underline{what} to \underline{whom}, \underline{when} and \underline{where}, etc. 
Semantic knowledge has been proved informative in many downstream NLP applications, such as question answering \citep{shen2007using, wang2015machine}, text summarization \citep{genest2011, khan2015}, and
machine translation \citep{liu2010semantic, gao2011corpus}.

Depending on how the semantic roles are defined, there are two forms of SRL in the community. 
The span-based SRL follows the manual annotations in the PropBank \citep{palmer2005proposition} and NomBank \citep{meyers2004nombank} and uses a continuous word span to be a semantic role. 
In contrast, the dependency-based SRL fulfills a role with a single word, which is usually the syntactic or semantic head of the manually annotated span \citep{surdeanu2008conll}.

\begin{figure}[!tp]
		\centering
		\includegraphics[scale=1.0]{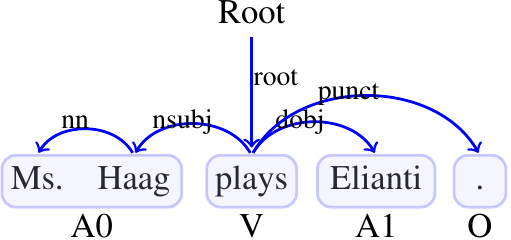}
		\caption{Example of dependency and SRL structures.}
		\label{fig:srl-example}
\end{figure}

This work follows the span-based formulation. 
Formally, given an input sentence $\bm{w}=w_1...w_n$ and a predicative word $\texttt{prd}=w_p$ ($1 \le p \le n$), the task is to recognize the semantic roles of $\texttt{prd}$ in the sentence, such as A0, A1, AM-ADV, etc. 
We denote the whole role set as $\mathcal{R}$. 
Each role corresponds to a word span of $w_j...w_k$ ($1 \le j \le k \le n$).  Taking Figure \ref{fig:srl-example} as an example,``Ms. Hag'' is the A0 role of the predicate ``plays''.

In the past few years, thanks to the success of deep learning, researchers has proposed effective neural-network-based models and improved SRL performance by large margins \citep{zhou2015end, he2017deep, tan2017deep}.
Unlike traditional discrete-feature-based approaches that make heavy use of syntactic features, recent deep-neural-network-based approaches are mostly in an end-to-end fashion and give little consideration of syntactic knowledge.

Intuitively, syntax is strongly correlative with semantics.  
Taking Figure \ref{fig:srl-example} as an example, 
the \textit{A0} role in the SRL structure is also the subject (marked by \textit{nsubj}) in the dependency tree, and the \textit{A1} role is the direct object (marked by \textit{dobj}). 
In fact, the semantic \textit{A0} or \textit{A1} argument of a verb predicate are usually the syntactic subject or object interchangeably according to the PropBank annotation guideline. 

In this work, we investigate several previous approaches for encoding syntactic trees, and make a thorough study on whether extra syntax-aware representations are beneficial for neural SRL models. The four approaches, Tree-GRU, Shortest Dependency Path (SDP), Tree-based Position Feature (TPF), and Pattern Embedding (PE), 
try to encode useful syntactic information in the input dependency tree from different perspectives. 
Then, we use the encoded syntax-aware representation vectors as extra input word representations, requiring little change of the architecture of the basic SRL model. 

For the base SRL model, we employ the recently proposed deep highway-BiLSTM model \citep{he2017deep}. 
Considering that the quality of the parsing results has great impact on the performance of syntax-aware SRL models, we employ the state-of-the-art biaffine parser to parse all the data in our work, which achieves 94.3\% labeled parsing accuracy on the WSJ test data \citep{dozat2016deep}.

We conduct our experiments on the benchmark CoNLL-2005 dataset, comparing our syntax-aware SRL approaches with a strong baseline with external word representations from ELMo. 
Detailed error analyses also give us more insights on the investigated approaches. 
The results show that, with the extra syntax-aware representations, our approach achieves new state-of-the-art 85.6 F1 (single model) and 86.6 F1 (ensemble) on the test set, outperforming the corresponding strong baselines with ELMo by 0.8 and 1.0, respectively, demonstrating the usefulness of syntactic knowledge.

\section{The Basic SRL Architecture}

Following previous works  \citep{zhou2015end,he2017deep,tan2017deep}, we also treat the task as a sequence labeling problem and try to find the  highest-scoring tag sequence $\hat{\bm{y}}$. 
\begin{equation}
\hat{\bm{y}} = \argmax_{\bm{y} \in \mathcal{Y}(\bm{w})} score(\bm{w}, \bm{y})
\end{equation}
where $y_i \in \mathcal{R'}$ is the tag of the $i$-th word $w_i$, and $\mathcal{Y}(\bm{w})$ is the set of all legal sequences. Please note that $\mathcal{R'} = (\{B, I\} \times \mathcal{R}) \cup \{O\}$. 

In order to compute $score(\bm{w}, \bm{y})$, which is the score of a tag sequence $\bm{y}$ for $\bm{w}$, we directly adopt the architecture of \cite{he2017deep}, which consists of the following four components, as illustrated in Figure \ref{fig:highway-example}. 

\subsubsection{The Input Layer} 

\begin{figure}
		\centering
		\includegraphics[scale=1.0]{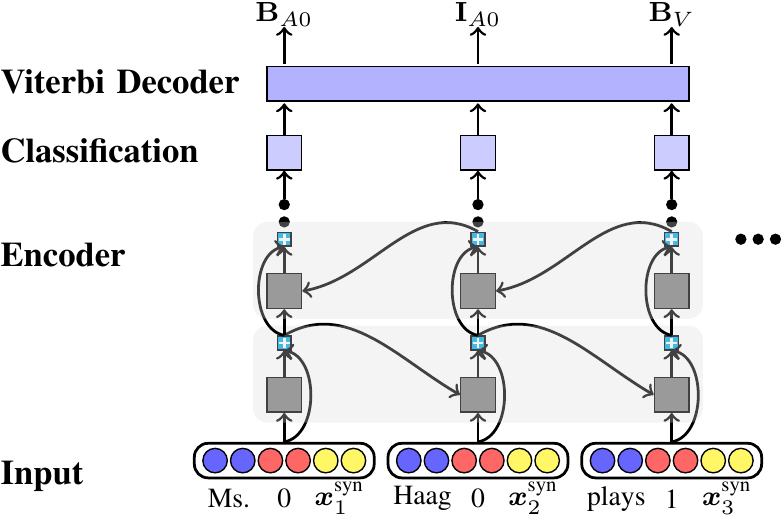}
		\caption{The basic SRL architecture.}
		\label{fig:highway-example}
\end{figure}

Given the sentence $\bm{w}=w_1...w_n$ and the predicate $\texttt{prd}=w_p$, the input of the network is the combination of the word embeddings and the predicate-indicator embeddings. Specifically, the input vector at the $i$-th time stamp is 
\begin{equation}
\bm{x}_i = \bm{emb}^{\texttt{word}}_{w_i} \oplus \bm{emb}^{\texttt{prd}}_{i==p}
\end{equation}
where the predicate-indicator embedding $\bm{emb}^{\texttt{prd}}_{0}$ is used for non-predicate positions and $\bm{emb}^{\texttt{prd}}_{1}$ is used for the $p$-th position, in order to distinguish the predicate word from other words, as shown in Figure \ref{fig:highway-example}.

With the predicate-indicator embedding, the encoder component can represent the sentence in a predicate-specific way, leading to superior performance \citep{zhou2015end,he2017deep,tan2017deep}. However, the side effect is that we need to separately encode the sentence for each predicate, dramatically slowing down training and evaluation.

\subsubsection{The BiLSTM Encoding Layer} 

Over the input layer, four stacking layers of BiLSTMs are applied to fully encode long-distance dependencies in the sentence and obtain the rich predicate-specific  token-level representations. 

Moreover, \cite{he2017deep} propose to use highway connections \citep{srivastava2015training, zhang2016highway} to alleviate the vanishing gradient problem,  improving the parsing performance by 2\% F1. 
As illustrated in Figure \ref{fig:highway-example}, the basic idea is to combine the input and output of an LSTM node in some way, and feed the combined result as  the final output of the node into the next LSTM layer and the next time-stamp of the same LSTM layer. 

We use the outputs of the final (top) backward LSTM layer as the representation of each word, denoted as $\bm{h}_i$.

\subsubsection{Classification Layer}

With the representation vector $\bm{h}_i$ of the word $w_i$, we employ a linear transformation and a softmax  operation to compute the probability distribution of different tags, denoted as $p(r|\bm{w}, i)$ ($r \in \mathcal{R}'$).

\subsubsection{Decoder}

With the local tag probabilities of each word, then the score of a tag sequence is 
\begin{equation}
score(\bm{w}, \bm{y}) = \sum_{i=1}^{n} \log p(y_i | \bm{w}, i)
\end{equation}

Finally, we employ the Viterbi algorithm to find the highest-scoring tag sequence and ensure the resulting sequence does not contain illegal tag transitions such as $y_{i-1}={B}_{A0}$ and $y_{i}={I}_{A1}$.

\section{The Syntax-aware SRL Approaches}

The previous section introduces the basic model architecture of SRL, 
and in this section, we will illustrate how to encode the syntactic features into a dense vector and use it as extra inputs. 
Intuitively, dependency syntax has a strong correlation with semantics. 
For instance, a subject of a verb in dependency trees usually corresponds to the agent or patient of the verb. 
Therefore, traditional discrete-feature based SRL approaches make heavy use of syntax-related features. 
In contrast, the state-of-the-art neural network based SRL models usually adopt the end-to-end framework without consulting the syntax.

This work tries to make a thorough investigation on whether integrating syntactic knowledge is beneficial for state-of-the-art neural network based SRL approaches. 
We investigate and compare four different approaches for encoding syntactic trees. 
\begin{itemize}
    \item The \emph{Tree-GRU} (tree-structured gated recurrent unit) 
    approach globally encodes an entire dependency tree once for all, and produces unified representations of all words in a predicate-independent way. 
    \item The \emph{SDP} (shortest dependency path) approach considers the shortest path from the focused word $w_i$ and the predicated $w_p$, and use the max-pooling of the syntactic labels in the path as the predicate-specific (different from Tree-GRU) syntax-aware representation of $w_i$. 
    \item The \emph{TPF} (tree position feature) approach considers the relative  positions of $w_i$ and $w_p$ to their common ancestor word in the parse tree, and use the embedding of such position features as the representation of $w_i$. 
    \item The \emph{PE} (pattern embedding) approach classifies the relationship  between $w_i$ and $w_p$ in the parse tree into several pre-defined patterns (or types), and use the embeddings of the pattern and a few dependency relations as the representation of $w_i$. 
\end{itemize}
The four approaches encode dependency trees from different perspectives and in different ways. 
Each approach produces a syntax-aware representation of the focused word. 

Formally, we denote these representations as $\bm{x}^{\texttt{syn}}_i$ for word $w_i$.
We treat these syntactic representations as the external model input, and concatenate them with the basic input $\bm{x}_i$.
In the following, we introduce the four approaches in detail. 

\subsection{The Tree-GRU Approach}

As a straightforward method, tree-stuctured recurrent neural network (Tree-RNN) \citep{tai2015improved, chen2017improved} can globally encode a parse tree and return syntax-aware representation vectors of all words in the sentence. 
Previous works have successfully employed Tree-RNN for exploiting syntactic parse trees for different tasks, such as sentiment classification \citep{tai2015improved}, relation extraction \citep{miwa2016end, feng2017joint} and machine translation \citep{chen2017improved, wang2018tag}.

\begin{figure}[tb] 
	\centering{\includegraphics[scale=1.0]{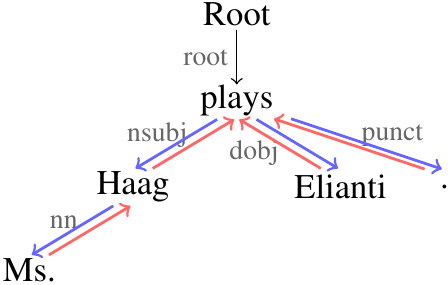}}
	\caption{Bi-directional (bottom-up and top-down) Tree-GRU.} 
	\label{fig:tree-gru}
\end{figure}

Following \cite{chen2017improved}, we employ a bi-directional Tree-GRU to encode the dependency tree of the input sentence, as illustrated in Figure \ref{fig:tree-gru}. 

The bottom-up Tree-GRU computes the representation vector $\bm{h}_i^\uparrow$ of the word $w_i$ based on its children, as marked by the red lines in Figure  \ref{fig:tree-gru}. 
The detailed equations are as follows. 
\begin{equation}
\begin{split}
&\bar{\bm{h}}_{i,L}^{\uparrow}  = \sum_{j \in lchild(i)}\bm{h}_j^{\uparrow}, \text{~~~~~~}
\bar{\bm{h}}_{i,R}^{\uparrow} = \sum_{k \in rchild(i)}\bm{h}_k^{\uparrow} \\
&\bm{r}_{i,L} = \sigma(\mathbf{W}^{rL}\bm{l}_i +  \mathbf{U}^{rL}\bar{\bm{h}}_{i,L}^{\uparrow} +
\mathbf{V}^{rL}\bar{\bm{h}}_{i,R}^{\uparrow}) \\
&\bm{r}_{i,R} = \sigma(\mathbf{W}^{rR}\bm{l}_i + \mathbf{U}^{rR}\bar{\bm{h}}_{i,L}^{\uparrow} +
\mathbf{V}^{rR}\bar{\bm{h}}_{i,R}^{\uparrow}) \\
&\bm{z}_{i,L} = \sigma(\mathbf{W}^{zL}\bm{l}_i + \mathbf{U}^{zL}\bar{\bm{h}}_{i,L}^{\uparrow} +
\mathbf{V}^{zL}\bar{\bm{h}}_{i,R}^{\uparrow}) \\
&\bm{z}_{i,R} = \sigma(\mathbf{W}^{zR}\bm{l}_i +  \mathbf{U}^{zR}\bar{\bm{h}}_{i,L}^{\uparrow} +
\mathbf{V}^{zR}\bar{\bm{h}}_{i,R}^{\uparrow}) \\
&\bm{z}_{i} = \sigma(\mathbf{W}^{z}\bm{l}_i + \mathbf{U}^{z}\bar{\bm{h}}_{i,L}^{\uparrow} +
\mathbf{V}^{z}\bar{\bm{h}}_{i,R}^{\uparrow}) \\
&\hat{\bm{h}}_i^{\uparrow} \text{$=$} \text{$\tanh$}\big(\mathbf{W}\bm{l}_i\text{$+$} \mathbf{U} (\bm{r}_{i,L} \text{$\odot$}\bar{\bm{h}}_{i,L}^{\uparrow}) \text{$+$}
\mathbf{V} (\bm{r}_{i,R} \text{$\odot$} \bar{\bm{h}}_{i,R}^{\uparrow} ) \big)\\
&\bm{h}_i^{\uparrow} = \bm{z}_{i,L} \odot \bar{\bm{h}}_{i,L}^{\uparrow} + \bm{z}_{i,R} \odot \bar{\bm{h}}_{i,R}^{\uparrow} + \bm{z}_{i} \odot \hat{\bm{h}}_i^{\uparrow},
\end{split}
\end{equation}
where $\texttt{lchild/rchild}(.)$ means the set of left/right-side children, $\bm{l}_i$ is the embedding of the syntactic label between $w_i$ and its head, 
and $\mathbf{W}$s, $\mathbf{U}$s and $\mathbf{V}$s are all model parameters.

Analogously, the top-down Tree-GRU computes the representation vector $\bm{h}_i^\downarrow$ of the word $w_i$ based on its parent node, as marked by the blue lines in Figure  \ref{fig:tree-gru}. 
We omit the equations for brevity. 

We use the concatenation of the two resulting hidden vectors as the  the final representation of the word $w_i$: 
\begin{equation} 
		\bm{x}^\texttt{syn}_i = \bm{h}_i^{\uparrow} \oplus \bm{h}_i^{\downarrow}
\label{eq:tree-gru-con}
\end{equation}

\subsection{The SDP Approach}

SDP-based features have been extensively used in traditional discrete-feature based approaches for exploiting syntactic knowledge in relation extraction.
The idea is to consider the shortest path of two focused words in the dependency tree, and extract path-related features as syntactic clues. 

\cite{xu2015classifying} first adapt SDP into the nerual network settings in the task of relation classification. 
They treat the SDP as two sub-paths from the two focused entity words to their lowest common  ancestor, and run two LSTMs respectively along the two sub-paths 
to obtain extra syntax-aware representations, leading to improved performance. 

\begin{figure}[tb] 
	\centering{\includegraphics[scale=1.0]{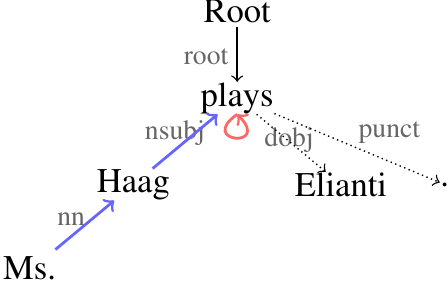}}
	\caption{The SDP approach: where the blue line marks the path from the focused word  ``\emph{Ms.}'' to the common ancestor ``\emph{plays}'', and the red line is the path from the predicate to the ancestor. } 
	\label{fig:sdp} 
\end{figure}

Directly employing LSTMs on SDPs leads to prohibitively efficiency problem in our scenario, because we have $O(n)$ paths given a sentence and a predicate. 
Therefore, we adopt the max pooling operation to obtain a representation vector over an SDP. Following \cite{xu2015classifying}, we divide the SDP into two parts in the position of the lowest  common ancestor, in order to distinguish the directions, as shown in Equation \ref{eq:sdp} and  Figure \ref{fig:sdp}. 
\begin{equation}\label{eq:sdp}
		\bm{x}^\texttt{syn}_i = \underset{j \in \texttt{path}(i,a)}{\mathrm{MaxPool}} (\bm{l}_j) \oplus \underset{k \in \texttt{path}(p,a)}{\mathrm{MaxPool}} (\bm{l}_k)
\end{equation}
where $\texttt{path}(i,a)$ is the set of all the words along the path from the focused word $w_i$ to the lowest common ancestor $w_a$, and $\texttt{path}(i,a)$ is for the path from the predicate $w_p$ to $w_a$; $\bm{l}_j$ is the embedding of the dependency relation label between $w_j$ and its parent.

\subsection{The TPF Approach} 

\cite{collobert2011natural} first propose position features for the task of SRL. 
The basic idea is to use the embedding of the distance (as discrete numbers) between the  predicate word and the focused word as extra inputs.

\begin{figure}[tb] 
	\centering{\includegraphics[scale=1.0]{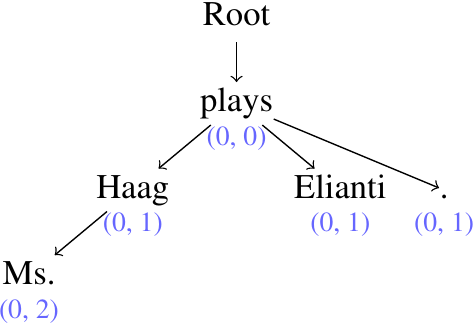}}
	\caption{Example of Tree-based Position Feature. The number-tuples with brackets are relative positions of TPF.}
	\label{fig:tpf2}
\end{figure}

In order to use syntactic trees, \cite{yang2016position} extend the position  features and propose the tree-based  position features (TPF) for the task of relation classification.

In this work, we directly adopt the TPF approach of \cite{yang2016position} for encoding syntactic knowledge.\footnote{\cite{yang2016position} propose two versions of TPF. We directly use the Tree-based Position Feature {2} due to its better performance.} 
Figure \ref{fig:tpf2} gives an example. 
The number pairs in the parentheses are the TPFs of the corresponding words. 
The first number means the distance from the predicate in concern to the lowest common ancestor, and the second number is the distance from the focused word to the ancestor. 
For instance, suppose ``Ms.'' is the focused word and ``plays'' is the predicate. 
Their lowest common ancestor is ``plays'', which is the predicate itself. 
There are 2 dependencies in the path from ``Ms.'' to ``plays''. 
Therefore, the TPF of ``Ms.'' is ``(0, 2)''.

Then, we embed the TPF of each word into a dense vector through a lookup operation, use it as the syntax-related representation. 
\begin{equation}
\begin{split}
	\bm{x}^\texttt{syn}_i = \bm{emb}_{f_i}^{\texttt{TPF}}
\end{split}
\end{equation}
where $f_i$ is the TPF of $w_i$.

\begin{figure}[tb] 
	\centering{\includegraphics[scale=1.0]{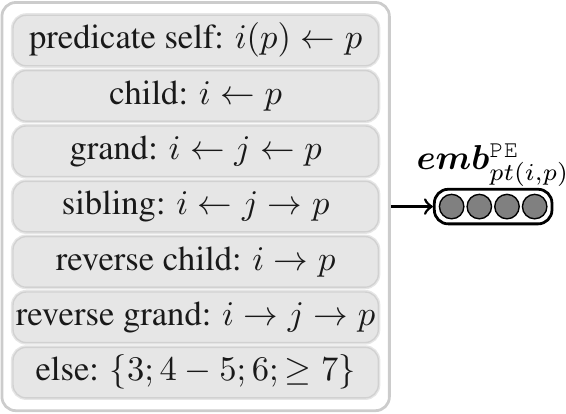}}
		\caption{Patterns used in this work. 
		}
	\label{fig:pattern}
\end{figure}

\subsection{The PE Approach} 

\cite{jiang2018supervised} propose the PE approach for the task of treebank conversion, which aims to convert a parse tree following the source-side annotation guideline into another tree following the target-side guideline. 
The basic idea is to classify the relationship between two given words in the source-side tree into several pre-defined patterns (or types), and use the embedding of the pattern and a few dependency relations as the source-side syntax representation.

In this work, we adopt their PE approach for encoding syntactic knowledge. 
Figure \ref{fig:pattern} lists the patterns used in this work. 
Given a focused word $w_i$ and a predicate $w_p$, we first decide the pattern type according to the structural relationship between $w_i$ and $w_p$, denoted as $pt(i,p)$. 

Taking $w_i=$``Ms.'' and $w_p=$ ``plays'' as an example, since ``Ms.'' is the grandchild of ``plays'', then the pattern is ``grandchild''. 
Then we embed $pt(i,p)$ into a dense vector $\bm{emb}^{\texttt{PE}}_{pt(i,p)}$.
We also use the embeddings of three highly related syntactic labels as extra representations, i.e., $\bm{l}_{i}$, $\bm{l}_{a}$, $\bm{l}_{p}$, where $\bm{l}_{a}$ is the syntactic label between the lowest common ancestor $w_a$ and its father. 
Then, the four embeddings are concatenated as $\bm{x}^\texttt{syn}_i$ to represent the structural information of $w_i$ and $w_p$ in a dependency tree. 
\begin{equation}
	\bm{x}^\texttt{syn}_i = \bm{emb}^\texttt{PE}_{pt(i,p)} \oplus \bm{l}_{i} \oplus \bm{l}_{a} \oplus \bm{l}_{p}
\end{equation}

\section{Experiments} 

\subsection{Settings} 
\subsubsection{Data} 
Following previous works, we adopt the CoNLL-2005 dataset with the standard data split: sections 02-22 of the Wall Street Journal (WSJ) corpus as the training dataset, section 24 as the development dataset, section 23 as the in-domain test dataset, sections 01-03 of the Brown corpus as the out-of-domain test dataset \citep{carreras2005introduction}. Table \ref{table:conll05-data} shows the statistics of the data.

\begin{table}[tb] 
	\setlength{\tabcolsep}{6pt} 
	\addtolength{\tabcolsep}{0.5pt}
	\begin{center}
			\begin{tabular}{l rrrr}
					\hline
					                         & Train     & Dev        & WSJ Test    & Brown Test    \\
					\hline
					{\#Sent}              &39,832    &1,346      &2,416  &426    \\
					{\#Tok}                 &950,028   &32,853     &56,684 &7,159  \\
					{\#Pred}             &90,750    &3,248      &5,267  &804    \\
					{\#Arg}              &239,858   &8,346      &14,077 &2,177  \\
					\hline
			\end{tabular}
			\caption{Data statistics of the CoNLL-2005.}
			\label{table:conll05-data}
	\end{center}
\end{table}

\setlength{\tabcolsep}{1.8pt}
\begin{table*}[tb]
	\renewcommand{\arraystretch}{1.2}
	\addtolength{\tabcolsep}{2pt}
	\begin{center}
		\begin{small}
			\begin{tabular}{l c  cccc cccc c}
				    \toprule
					  	&	& \multicolumn{4}{c}{WSJ Test} & \multicolumn{4}{c}{Brown Test} & Combined  \\ 
							\cmidrule(lr){3-6}		\cmidrule(lr){7-10}	\cmidrule(lr){11-11}
					&Methods     &P &R &F1 &Comp.           &P &R &F1 &Comp. &F1 \\ \hline
					\hline
					\multirow{6}{*}{Single}
					& Baseline \citep{he2017deep}       &83.1 &83.0 &83.1 &64.3   &72.9  &71.4  &72.1  &44.8  &81.6 \\
					&{Baseline} (Our re-impl) 	&83.4    &83.0    &83.2    &64.9      &72.3   &70.8 &71.6 &44.3  &81.6     \\
					&{Tree-GRU}			           &83.9 &83.6	&83.8 &65.2   &72.9	 &71.1	&72.3  &44.9  &82.2 \\
					&{SDP}				          &84.2 &\bf{83.9}	&84.0 &\bf{65.9}   &\bf{74.0}	 &72.0	&\bf{73.0}  &45.2  &\bf{82.6} \\
					&{TPF}			                &\bf{84.3} &83.8	&\bf{84.1} &\bf{65.9}   &73.7	 &72.0	&72.9  &45.5  &\bf{82.6} \\
					&{PE}                   		&83.7 &83.8 &83.8 &65.3   &73.4  &\bf{72.5}	&\bf{73.0}  &\bf{46.1}  &82.3 \\
					\hline
					\hline
					\multirow{5}{*}{Single+ELMo}
					&{Baseline}  	&86.3    &86.2    &86.3    &69.4      &75.2   &74.3 &74.7 &48.1  &84.8     \\
					&{Tree-GRU} &86.2	&86.2	&86.2	&68.9	&77.9	&75.6	&76.7	&50.8 &84.9 \\
					&{SDP} &86.9 &86.7 &86.8 &70.2 &\bf{78.0} &76.3 &\bf{77.1} &\bf{52.4} &85.5 \\
					&{TPF}			             &\bf{87.}0 &\bf{86.8}	&\bf{86.9} &\bf{70.4}   &77.6	 &75.9	&76.8  &51.6  &\bf{85.6} \\
					&{PE} &86.5	&86.3  &86.4	&69.6 	&77.4	&\bf{76.4}	&76.9	&51.5 &85.1 \\
					\hline
					\hline
					\multirow{3}{*}{Ensemble}
					&$5 \times$ Baseline \citep{he2017deep} &85.0 &84.3 &84.6 &66.5  
					&74.9 &72.4 &73.6 &46.5   &83.2 \\
					&$5 \times$ Baseline (Our re-impl)     &84.6 &84.0 &84.3 &66.4  &74.9 &72.1 &73.5 &46.1         &82.9 \\
					&{$5 \times$ TPF}			     	&85.6  &85.0 &85.3 &68.1   &75.9  &73.4  &74.8  &48.1  &83.9   \\
					&{4 Syntax-aware Methods}		       &\bf{85.8} &\bf{85.5} &\bf{85.6} &\bf{68.7}	 &\bf{76.3}&\bf{74.5}&\bf{75.4}  &\bf{49.3}  &\bf{84.3}\\
					\hline
					\hline
					\multirow{4}{*}{Ensemble+ELMo}
					&{$5 \times$ Baseline} &87.2 &86.8 &87.0 &70.8      &77.7 &75.8 &76.7 &50.4 &85.6 \\
					&{$5 \times$ TPF}	     	&87.5  &87.0 &87.3 &71.1   &78.6  &76.5  &77.5  &52.5  &86.0   \\
					&{4 Syntax-aware Methods} &\bf{88.0} &\bf{87.6} &\bf{87.8} &\bf{72.2} &\bf{79.7} &\bf{78.0} &\bf{78.8} &\bf{53.2} &\bf{86.6} \\
					\bottomrule
			\end{tabular}
				\caption{Comparison with baseline model and all our syntax-aware methods on the CoNLL-2005 dataset. We report the results in precision (P), recall (R), F1 and percentage of completely correct predicates (Comp.). } 
			\label{table:conll05-all}
		\end{small}
	\end{center}
\end{table*} 

\subsubsection{Dependency Parsing}
In recent years, neural network based dependency parsing has achieved significant progress. We adopt the state-of-the-art biaffine parser proposed by \cite{dozat2016deep} in this work. 
We use the original phrase-structure Penn Treebank (PTB) data to produce the dependency structures for the CoNLL-2005 SRL data. 
Following standard practice in the dependency parsing community, the phrase-structure trees are converted into Stanford dependencies  using the Stanford Parser v3.3.0\footnote{\scriptsize{\url{https://nlp.stanford.edu/software/lex-parser.html}}}. 
Since the biaffine parser needs part-of-speech (POS) tags as inputs, we use an in-house CRF-based POS tagger to produce automatic POS tags on all the data. 
After training, the biaffine parser achieves 94.3\% parsing accuracy (LAS) on the WSJ test dataset. 
Additionally, we use the 5-way jackknifing to obtain the automatic POS tagging and dependency parsing results of the training data.

\subsubsection{ELMo} 

\cite{Peters:2018} recently propose to produce contextualized word representations (ELMo) with an unsupervised language model learning objective, and show that simply using the learned external word representations as extra inputs can effectively boost performance of a variety of tasks, including SRL. 
To further investigate the effectiveness of our syntax-aware methods, we build a stronger baseline with the ELMo representations as extra input.  

\subsubsection{Evaluation} 
We adopt the official script provided by 
CoNLL-2005\footnote{\scriptsize{\url{http://www.cs.upc.edu/~srlconll/st05/st05.html}}} for evaluation.  
We conduct significance test using the Dan Bikel's randomized parsing evaluation
comparer.

\subsubsection{Implementation} 
We implement the baseline and all the syntax-aware methods with Pytorch 0.3.0\footnote{\scriptsize{\url{github.com/KiroSummer/Syntax-aware-Neural-SRL}}}.

\subsubsection{Initialization and Hyper-parameters} 
For the parameter settings, we mostly follow the work of \cite{he2017deep}. 
We adopt the Adadelta optimizer with learning rate $\rho = 0.95$ and $\epsilon = 1e-6$, use a batchsize of $80$, and clip gradients with norm larger than $1.0$. 
All the embedding dimensions (word, predicate-indicator, syntactic label, pattern, and TPF) are set to $100$. 
All models are trained for $500$ iterations on the trained data and select the best iteration that has the peak performance on the dev data.

\subsection{Main Results} \label{Results}

Table \ref{table:conll05-all} shows the main results of different  approaches on CoNLL-2005 dataset. 
The results are presented in four major rows. 
For the ensemble of ``5 $\times$ baseline'' and ``5 $\times$ TPF'', we randomly sample 4/5 of the training data and train one model at each time.
For the ensemble of 4 syntax-aware approaches, each model is trained on the whole training data.

\textbf{Results of single models} are shown in the first major row. 
First, our re-implemented baseline of \cite{he2017deep} achieves nearly the same results with those reported in  their  paper.
Second, the four syntax-aware approaches are similarly  effective and can improve the performance by $0.6 \sim 1.0$  in F1 (combined). All the improvements are statistically significant ($p < 0.001$). 
Third, we find that the Tree-GRU approach is slightly inferior to the other three. The possible reason is that Tree-GRU produces unified representations for the  sentence without specific considerations of the given predicate, whereas all other three approaches derive predicate-specific representations.

\textbf{Results of single models with ELMo} are shown in the second major row, in which each single model is enhanced using the ELMo representations as extra inputs. 
We can see that ELMo representations brings substantial improvements over the corresponding baselines by $2.7 \sim 3.2$ in F1 (combined). 
Compared with the stronger baseline w/ ELMo, SDP, TPF, and PE w/ ELMo still achieve absolute and significant ($p < 0.001$) improvement of $0.7$, $0.8$, and $0.3$ in F1  (combined), respectively. 
Tree-GRU w/ ELMo increases F1 by $2.0$ on the  out-of-domain Brown test data, but decreases F1 by $0.1$ on the in-domain WSJ test data, leading to an overall improvement of $0.1$ on combined F1 ($p > 0.05$). 
Similarly to the results in the first major row, this again indicates that the predicate-independent Tree-GRU approach may not be suitable for syntax encoding in our SRL task, especially when the baseline is strong.

\vspace{+3pt}
\textbf{Results of ensemble models} are shown in the third major row. 
Compared with the baseline single model, the baseline ensemble approaches increase F1 (combined) by $1.3$. 
The F1 score of the ``5 $\times$ Baseline'' of \citep{he2017deep} is 83.2\%, whereas our re-implemented ``5 $\times$ Baseline'' achieves 82.9\% F1.
We guess the 0.3 gap may be caused by the random factors in 5-fold data split and parameter initialization.
In our preliminary experiments, we have run the single model of \cite{he2017deep} for several times using different random seeds for parameter initialization, and found about $0.1 \sim 0.3$ F1 variation.
The ensemble of five TPF models further improves F1 (combined) over the ensemble of five baselines by $1.0$ ($p < 0.001$). 
We choose the TPF  method as a case study since it  consistently achieves best performances in all scenarios.
The ensemble of the four syntax-aware methods achieves the best performance in this scenario, outperforming the TPF ensemble by $0.4$ F1. 
This indicates that the four syntax-aware approaches are discrepant and thus more complementary in the ensemble scenario than using a single method. 

\vspace{+3pt}
\textbf{Results of ensemble models with ELMo} are shown in the bottom major row, where each single model is enhanced with ELMo representations before the ensemble operation. 
Again, using ELMo representations as extra input greatly improves F1 (combined) by $2.1 \sim 2.7$ over the corresponding ensemble methods in the third major row. 
Similar to the above findings, the ensemble of five TPF with ELMo improve F1 (combined) by $0.4$ ($p < 0.001$) over the ensemble of five baselines with ELMo, 
and 
the ensemble of the four syntax-aware methods with ELMo further increases F1 (combined) by $0.6$. 

\emph{Overall, we can conclude that the syntax-aware approaches can consistently improve SRL performance. }

\subsection{Comparison with Previous Works}

\setlength{\tabcolsep}{1.8pt}
\begin{table}[tb]
	\renewcommand{\arraystretch}{1.2}
	\addtolength{\tabcolsep}{0pt}
	\begin{center}
	        \begin{small}
			\begin{tabular}{cc ccc}
				    \toprule
					           &             & WSJ      &Brown     &Combi \\
					\hline
					\centering
					\multirow{5}{*}{Single}
					& TPF w/ ELMo     &\bf{86.9}       &\bf{76.8}       &\bf{85.6}    \\
					& TPF            &84.1       &72.9       &82.6       \\
					&\cite{strubell2018linguistically} &83.9  &72.6  &-         \\
					&\cite{he2017deep}      &83.1       &72.1       &81.6       \\
					&\cite{tan2017deep}     &{84.8}  &{74.1}  &{83.4}  \\
					\hline
					\multirow{5}{*}{Ensemble}
					    & 4 Syn-a Methods w/ ELMo        &\bf{87.8}  &\bf{78.8}    &\bf{86.6}   \\
					    & 4 Syntax-aware Methods                 &85.6    &75.4     &84.3   \\
					    &\cite{he2017deep}   &{84.6}  &73.6     &83.2   \\
					    &\cite{tan2017deep}  &86.1    &74.8     &84.6   \\
					    &\cite{fitzgerald2015semantic} &80.3 &72.2 &-   \\
					\bottomrule
			\end{tabular}
			\end{small}
				\caption{Comparison with previous results. } 
			\label{table:conll05-compare}
	\end{center}
\end{table} 

Table \ref{table:conll05-compare} compare our approaches with previous works. 
In the single-model scenario, the TPF approach outperforms both  \cite{he2017deep} and \cite{strubell2018linguistically}, and is only inferior to \cite{tan2017deep}, which is based on the recently proposed deep self-attention encoder \citep{vaswani2017attention}. 
Using ELMo representations promotes our results as the state-of-the-art. 
We discuss the work of \cite{strubell2018linguistically} in the related works section, which also makes use of syntactic information. 

In the ensemble scenario, the findings are similar, and the ensemble of four syntax-aware approaches with ELMo reaches new state-of-the-art performances.

\subsection{Analysis}

\begin{table}[tb]
	\addtolength{\tabcolsep}{2pt}
	\begin{center}
			\begin{tabular}{c c c c c}
				\toprule
					Methods  	& \multicolumn{1}{c}{Devel} & \multicolumn{1}{c}{WSJ} & \multicolumn{1}{c}{Brown} & Combined  \\ 
					\hline
					{Baseline}  			 	&81.5			&83.2     	  	&71.6   	&81.6     \\
					{TPF}			         	&82.5   	    &84.1    		&72.9    	&82.6 \\
					{TPF-Gold}			 	&88.4       	&89.6    		&79.8    	&88.3 \\
					\hline
					{Baseline w/ ELMo} &85.5 &86.3 &74.7 &84.8 \\
					{TPF w/ ELMo} &85.3 &86.9 &76.8 &85.6 \\
					{TPF-Gold w/ ELMo} &89.8  &91.1  &82.2  &89.9 \\ 
					\bottomrule
				\end{tabular}
				\caption{Upper-bound analysis of the TPF Method.} 
			\label{table:conll05-upper-bound}
	\end{center}
\end{table}

In this section, we conduct detailed analysis to better understand the improvements introduced by the syntax-aware approaches. We use the TPF approach as a case study since it consistently achieves best performances in all scenarios. We sincerely thank Luheng He for the kind sharing of her  analysis scripts. 

\begin{figure}[!tb]
		\centering
		\includegraphics[scale=0.8]{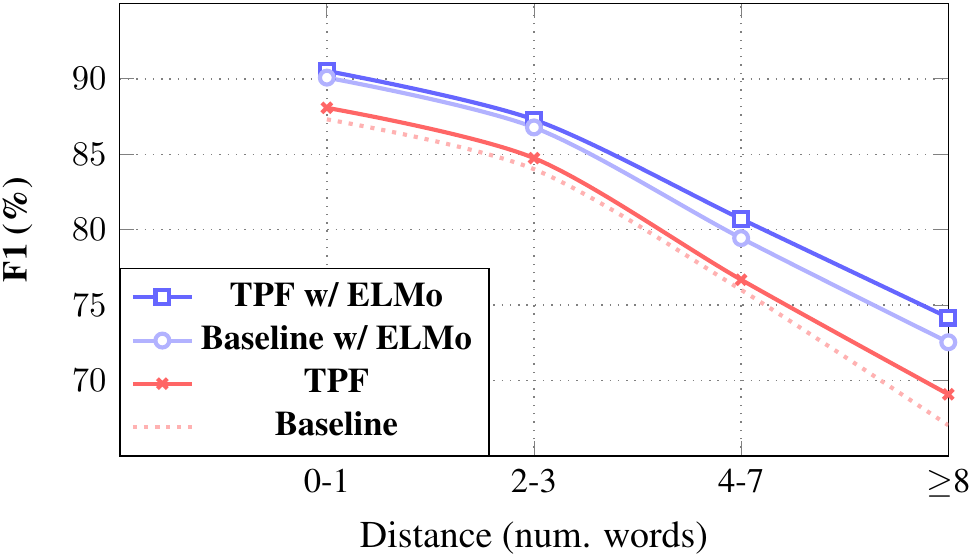}
		\caption{F1 regarding surface distance between arguments and predicates.} 
		\label{fig:distance-analysis}
\end{figure}

\begin{figure}[!tb]
		\centering
		\includegraphics[scale=0.8]{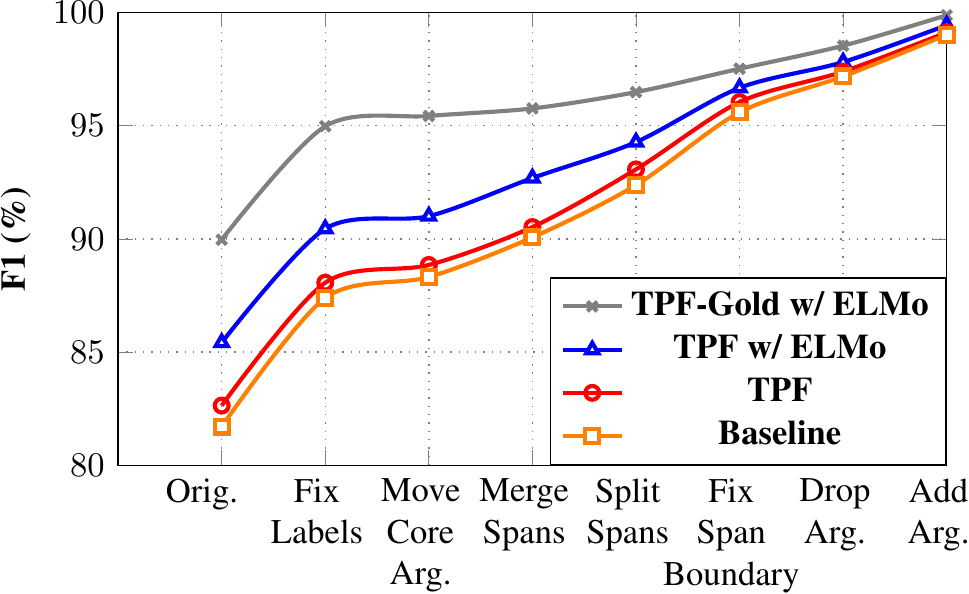}
		\caption{Performance of CoNLL-2005 models after performing oracle transformations.}
		\label{fig:oracle-transformation}
\end{figure}

\subsubsection{Using Gold-standard Syntax}

To understand the upper-bound performance of the syntax-aware approaches, we use the gold-standard dependency trees as the input and apply the TPF approach. 
Table \ref{table:conll05-upper-bound} shows the results. 
The TPF method with gold-standard parse trees brings a very large improvement of $6.7$ in F1 (combined) over the baseline without using ELMo, and $5.1$ when using ELMo. 
This shows the usefulness and great potential of syntax-aware SRL approaches. 

\subsubsection{Long-distance Dependencies}

To analyze the effect of syntactic information regarding to the distances between arguments and  predicates, 
we compute and report F1 scores of different sets of arguments according to their distances from  predicates, as shown in Figure \ref{fig:distance-analysis}. 
It is clear that larger improvements are obtained for arguments longer-distance arguments, in both scenarios of with or without ELMo representations. 
This demonstrates that \emph{syntactic knowledge effectively captures long-distance dependencies and thus is most beneficial for arguments that are far away from predicates.}

\subsubsection{Error Type Breakdown} 

In order to understand error distribution of different approaches in terms of different types of mistakes, we follow the work of \cite{he2017deep} and employ a set of oracle transformations on the system outputs to observe the relative F1 improvements by fixing various prediction errors incrementally. 
``Orig'' corresponds to the F1 scores of the original model outputs. 
First, we fix the label errors in the model outputs and the fixed results are shown by ``Fix Labels''. 
Specifically, if we find a predicted span matches a gold-standard one but has a wrong label, then we assign the correct label to the span in the model outputs. 
Then, based on the results of ``Fix Labels'', we perform ``Move Core Arg.''. 
If a span is labeled as a core argument (i.e., A0-A5), but the boundaries are wrong, then we move the span to its correct boundaries.
Third, based on the results of ``Move Core Arg.'', we perform ``Merge Spans''. 
If two predicted spans can be merged to match a gold-standard span, then we do so and assign the correct label.
Fourth, we preform ``Split Spans’’.
If a predicted span can be split into two gold-standard spans, then we do so and assign correct labels to them.
Fifth, if a predicted span's label matches an overlapping gold span, we perform ``Fix Span Boundary'' to correct its boundary.
Sixth, we perform ``Drop Arg.'' to drop the predicted arguments that doesn't overlap with any gold spans.
Finally, if a gold argument doesn't overlap with any predicted spans, we perform ``Add Arg.''
Figure \ref{fig:oracle-transformation} shows the results, from which we can see that 1) different approaches have similar error distributions, among which labeling errors account for the largest proportion, and span boundary errors (split, merge, boundary) also have a large share; 2) using automatic parse trees leads to consistent improvements over all error types; 3) using ELMo representations also consistently improves performance by large margin; 4) using gold parse trees can effectively resolve almost all span boundary (including merge and split) errors.

\section{Related Work}\label{sec:rel-work}

Traditional discrete-feature based SRL models make heavy use of syntactic information \citep{swanson2006comparison, punyakanok2008importance}. 
With the rapid development of deep learning in NLP, researchers propose several simple yet effective end-to-end neural network models with little consideration of syntactic knowledge \citep{zhou2015end, he2017deep, tan2017deep}.

Meanwhile, inspired by the success of syntactic features in traditional SRL approaches, researchers also try to enhance neural network based SRL approaches by syntax. 
\cite{he2017deep} show that large improvement can be achieved by using gold-standard constituent trees as rule-based constraints during viterbi decoding.
\cite{strubell2018linguistically} propose a  syntactically-informed SRL approach based on the  self-attention mechanism. The key idea is introduce an auxiliary training objective that encourages one attention head to attend to its syntactic head word. They also use multi-task learning on POS tagging, dependency parsing, and SRL to obtain better encoding of the input sentence.  
We make comparison with their results in Table  \ref{table:conll05-compare}.  
\cite{swayamdipta2018syntactic} propose a multi-task learning framework to incorporate constituent parsing  loss into other semantic-related tasks such as SRL and coreference resolution. 
They report +0.8 F1 improvement over their baseline SRL model.
Different from \cite{swayamdipta2018syntactic}, our work focuses on dependency parsing and try to explicitly encode parse outputs to help SRL.

The dependency-based SRL task is started in 
CoNLL-2008 shared task \citep{surdeanu2008conll}, which aims to jointly tackle syntactic and semantic dependencies.
There are also a few recent works on exploiting dependency trees for neural dependency-based SRL. 
\cite{roth2016neural} proposes an effective approach to obtain dependency path embeddings and uses them as extra features in traditional discrete-feature based SRL. 
\cite{marcheggiani2017encoding} propose a dependency tree encoder based on a graph convolutional network (GCN), which has a similar function as Tree-GRU, and stack the tree encoder over a sentence encoder based on multi-layer BiLSTMs. 
\cite{he2018syntax} exploit dependency trees for dependency-based SRL by 1) using dependency label  embeddings as extra inputs, and 2) employing a tree-based k-th order algorithm for argument pruning. 
\cite{cai2018full} propose a strong end-to-end neural approach for the dependency-based SRL based on deep BiLSTM encoding and Biaffine attention. They use dependency trees for argument pruning and find no improvement over the syntax-agnostic counterpart. 

\section{Conclusions}

This paper makes a thorough investigation and comparison of four different approaches, i.e., Tree-GRU, SDP, TPF, and PE, for exploiting syntactic knowledge for neural SRL. 
The experimental results show that syntax is consistently helpful to improve SRL performance, even when the models are enhanced  with external ELMo representations. 
By utilizing both ELMo and syntax-aware representations, our final models achieve new state-of-the-art 
performance in both single and ensemble scenarios on  the benchmark  CoNLL-2005 dataset. 
Detailed analyses show that syntax helps the most on  arguments that are far away from predicates, due to the long-distance dependencies captured by syntactic trees.  
Moreover, there is still a large performance gap between using gold-standard and automatic parse trees, 
indicating that there is still large room for further research on syntax-aware SRL.

\section{Acknowledgments}
We thank our anonymous reviewers for their helpful comments.
This work was supported by National Natural Science Foundation of China (Grant No. 61525205, 61876116,  61432013), and was partially supported by the joint research
project of Alibaba and Soochow University.

\bibliography{aaai19}
\bibliographystyle{aaai}
\end{document}